# An AI-driven framework for rapid and localized optimizations of urban open spaces

Pegah Eshraghi [1], Arman Nikkhah Dehnavi [2], Maedeh Mirdamadi [3], Riccardo Talami [4 5*], Zahra-Sadat Zomorodian [6]

[1] Department of Construction, Shahid Beheshti University (SBU), Tehran, Iran- Email: p_eshraghi@sbu.ac.ir,

[2] Klesse College of Engineering and Integrated Design, The University of Texas at San Antonio, USA-
Email: arman.nikkhahdehnavi@utsa.edu,

[3] Department of Construction, Shahid Beheshti University (SBU), Tehran, Iran-
Email: mirdamadi.maedeh@gmail.com,

[4] Department of Architecture, College of Design and Engineering, National University of Singapore, 117566, Singapore.

[5] Singapore-ETH Centre, Future Resilient Systems II, 1 Create Way, 138602, Singapore
Email: rtalami@nus.edu.sg,

[6] Department of Construction, Shahid Beheshti University (SBU), Tehran, Iran- Email: z_zomorodian@sbu.ac.ir

## Abstract

As urbanization accelerates, open spaces are increasingly recognized for their role in enhancing sustainability and well-being, yet they remain underexplored compared to built spaces. This study introduces an AI-driven framework that integrates machine learning models (MLMs) and explainable AI techniques to optimize Sky View Factor (SVF) and visibility—key spatial metrics influencing thermal comfort and perceived safety in urban spaces. Unlike global optimization methods, which are computationally intensive and impractical for localized adjustments, this framework supports incremental design improvements with lower computational costs and greater flexibility. The framework employs SHapley Adaptive Explanations (SHAP) to analyze feature importance and Counterfactual Explanations (CFXs) to propose minimal design changes. Simulations tested five MLMs, identifying XGBoost as the most accurate, with building width, park area, and heights of surrounding buildings as critical for SVF, and distances from southern buildings as key for visibility. Compared to Genetic Algorithms, which required approximately 15–30 minutes across 3–4 generations to converge, the tested CFX approach achieved optimized results in 1 minute with a 5% RMSE error, demonstrating significantly faster performance and suitability for scalable retrofitting strategies. This interpretable and computationally efficient framework advances urban performance optimization, providing data-driven insights and practical retrofitting solutions for enhancing usability and environmental quality across diverse urban contexts.

## 1. Introduction

As urbanization accelerates globally, cities face growing challenges in achieving sustainability, livability, and resilience. Balancing built and open spaces is essential for creating healthier, more inclusive, and environmentally responsive communities (Andreucci *et al.*, 2021; Johari *et al.*, 2020). While buildings and their energy performance have been widely studied (Toutou *et al.*,2018; Yan *et al.*,2022), open spaces—such as parks and green corridors— remain underexplored, despite their vital role in mitigating urban heat islands (UHIs), improving air quality, regulating microclimates, and promoting well-being (Chen *et al.*, 2012s; Olszewska-Guizzo *et al.*, 2022). Open spaces also support social interaction, recreation, and environmental sustainability (Bourbia and Boucheriba, 2010), but their usability and functionality depend heavily on the surrounding urban morphology, including building height, density, orientation, and proximity (Stamps, 1999s; Kaplan and Kaplan, 1989). These factors influence solar access, visibility, and thermal comfort, shaping perceptions of safety, accessibility, and usability (Zheng *et al.*,2023; Türkseven Doğrusoy and Zengel, 2017).

Among spatial metrics, Sky View Factor (SVF) and visibility have emerged as key indicators for evaluating openness and visual access in urban environments. SVF quantifies sky openness, serving as a measure of thermal comfort, solar exposure, and microclimatic regulation (Miao *et al.*, 2020; Wang and Akbari, 2014), while visibility affects safety perceptions, navigation, and social interactions, making it essential for urban design (Türkseven Doğrusoy and Zengel, 2017).

Despite the importance of spatial metrics like SVF and visibility, existing tools often emphasize the evaluation of built environments rather than the dynamics of open spaces. While SVF and visibility are widely recognized as indicators of openness, light availability, and perceived safety, most approaches focus solely on measuring these metrics without proposing adjustments to improve usability or environmental performance (Jim and Chen, 2006; Zavadskas *et al.*,2019; Sezavar *et al.*,2023). Furthermore, several studies have optimized SVF through simulation and parametric modeling techniques (Geng *et al.* 2024; Xu *et al.*, 2019). However, these approaches primarily rely on global search algorithms that are computationally demanding and require large-scale redesigns, making them impractical for localized adjustments (Toutou *et al.*, 2018; Xu *et al.*, 2019). Given that urban open spaces are often embedded within existing contexts, large-scale modifications may not always be feasible. Instead, localized interventions are more suitable, preserving existing structures while enhancing performance. This underscores the need for optimization approaches that balance computational efficiency, flexibility, and interpretability, enabling targeted modifications without requiring complete transformations, making them practical for retrofitting as well as new designs.

Counterfactual Explanations (CFXs) address this need by focusing on minimal, actionable changes to optimize spatial performance metrics. Unlike traditional optimization approaches—such as genetic algorithms (GA) and multi-objective methods like Non-Dominated Sorting Genetic Algorithm (NSGA-II) which emphasize global solutions (Saad and Araji, 2021)—CFXs prioritize localized adjustments, enabling incremental changes with lower computational costs, making them ideal for retrofitting and urban design optimizations (Tsirtsis and Gomez-rodriguez, 2020; Regenwetter *et al.,*2023).

To address these challenges, this study introduces an AI-driven framework that combines machine learning models (MLMs) with explainable AI techniques for interpretable and actionable optimization. To enhance interpretability, the framework leverages SHapley Adaptive ExPlanations (SHAP) for feature importance analysis (Zhong *et al.*, 2022) and CFXs to suggest minimal design changes to enhance SVF and visibility (Tsirtsis and Gomez-rodriguez, 2020). By integrating these methods, this approach provides an efficient, scalable, and transparent framework for optimizing urban open spaces.

## Aims, Objectives, and Contributions

This study develops an AI-driven framework that integrates SHAP and CFXs to evaluate and optimize spatial metrics in urban open spaces. Unlike traditional approaches that emphasize energy metrics and require global redesigns, this framework focuses on localized interventions, making it scalable, interpretable, and practical for urban retrofitting and design optimization. The focus on SVF and visibility highlights their location-independent nature, as they are determined solely by urban morphology rather than climate or geographical conditions. This characteristic ensures the generalizability of findings across diverse urban contexts. The framework's flexibility and scalability are demonstrated using Tehran, Iran as a case study, leveraging its varied urban configurations, including high-density cores and low-density areas, to test its adaptability. The primary aim of this study is to combine MLMs with explainable AI techniques to optimize spatial metrics in urban design. Key objectives include:

- Evaluating the influence of urban morphology parameters on SVF and visibility to assess their role in usability and environmental performance.

- Developing predictive models using MLMs to estimate SVF and visibility with high accuracy and computational efficiency.

- Employ SHAP to explain the importance of spatial features, enabling transparent analysis.

- Optimize spatial configurations through CFXs, proposing minimal changes (e.g., building heights and distances) without requiring global redesigns.

This research makes several key contributions to urban planning and design. It introduces a novel AI-based framework that integrates SHAP and CFXs to optimize SVF and visibility, addressing gaps in traditional methods. By enabling incremental optimizations, it reduces computational costs and supports data-driven decision-making. Tested in Tehran's diverse urban configurations, the framework demonstrates its flexibility, scalability, and generalizability across different urban contexts. It establishes a new standard for balancing built and open environments, showing how AI techniques can inform urban retrofitting and design flexibility.

The paper is organized as follows: Section 2 reviews prior research; Section 3 outlines methodology; Section 4 discusses findings; and Section 5 concludes with planning implications.

## 2. Literature review

Urban parks are essential for enhancing health, quality of life, and urban aesthetics while mitigating negative impacts like UHIs and pollution (Chen *et al.*, 2012). Several studies have explored how parks contribute to physical and mental well-being through recreation and social interaction (Chiesura, 2004; Luymes and Tamminga, 1995). However, the usability and functionality of parks are not solely determined by their internal layouts but also by their spatial relationships with surrounding buildings.

Building density, height, and spatial arrangements have been shown to influence safety, accessibility, and comfort (Stamps, 1999; Kaplan and Kaplan, 1989). For instance, Kothencz and Blaschke (2017) demonstrated that building density impacts perceptions of accessibility and safety. Despite these findings, most studies analyze these factors in isolation, failing to consider their interactions in shaping visibility and openness within open spaces. While some research investigates the effects of vegetation, spatial enclosure, and layout on safety and comfort (Jim and Chen 2006; Sezavar *et al.*2023), relatively few studies focus on the specific influence of building height and density on SVF and visibility—key factors that directly impact light exposure, perceived openness, and user safety (Lyu *et al.*, 2019; Miao *et al.*, 2020).

SVF, first introduced by Oke (1981), evaluates thermal comfort, microclimates, and solar exposure (Lyu *et al.*,2019; Wang and Akbari, 2014). While traditional methods like mathematical models and satellite imagery (Watson and Johnson, 1987; Chang *et al.*, 2020; Shams Amiri *et al.,* 2023) are reliable, they often lack scalability and computational efficiency for larger urban scenarios. Existing approaches to optimizing SVF, such as parametric modeling and simulation-based methods (Geng *et al.* 2024; Xu *et al.*, 2019), frequently rely on global optimization strategies that require large-scale redesigns of urban configurations. These methods are computationally intensive and less suitable for retrofitting existing spaces or making localized adjustments.

Optimization techniques, particularly GAs and their variants, such as NSGA-II, have been widely used to explore urban design alternatives and optimize spatial configurations (Saad and Araji, 2021). Tools like Galapagos and Octopus leverage these algorithms to generate design solutions (Jiaweiyao et al., 2021). However, their reliance on global search methods and high computational costs often limit their practicality, especially for localized adjustments (Toutou 2018; Xu et al., 2019). Although problem decomposition methods have been proposed to improve computational efficiency and scalability (Talami *et al.*, 2020) their ability to handle localized interventions in urban optimization remains underexplored. While NSGA-II effectively diversifies solutions and improves convergence speed (Liu *et al.*, 2023), its time-intensive computational process restricts flexibility in applications requiring rapid and targeted adjustments (Saad and Araji, 2021).

To overcome these limitations, recent advances in MLMs and Explainable AI techniques offer new pathways for scalable and interpretable optimization. MLMs reduce computational processing times while maintaining accuracy, making them suitable for spatial metric analysis (Slack *et al.,*2020). Tools like SHAP highlight influential features in predicting metrics such as energy consumption (Dinmohammadi *et al.,* 2023; Sun *et al.*, 2022), thermal comfort (Zhang *et al.*, 2021; Yang *et al.*, 2022), and UHI (Yu *et al.*, 2020; McCarty *et al.,* 2021), as well as in the development of tools for building performance analysis (Eshraghi *et al.*, 2024; Shen and Pan, 2023). However, SHAP focuses on interpretation rather than suggesting design changes, limiting its application to optimization tasks.

To complement SHAP, CFXs propose actionable modifications for improving spatial metrics. CFXs, widely used in economics and medicine (Tsirtsis and Gomez-rodriguez, 2020; Regenwetter *et al.*, 2023), offer potential in urban planning for optimizing spatial metrics by providing actionable recommendations. Unlike global search algorithms, which propose complete redesigns, CFXs deliver incremental changes, such as modifying building heights or distances, making them practical for both new developments and retrofitting existing layouts. This capability is particularly useful in urban contexts, where flexibility and minimal disruption are often priorities.

## 3. Method

The research methodology comprises four phases: (1) Performance Simulation, (2) Building the Predictive Models, (3) Feature Importance Assessment, (4) CFXs for Scenario-specific Feature Evaluation, and (5) Validation by Grasshopper and GA optimization through the Galapagos plug-in (Figure 1). The subsequent sections detail each phase.

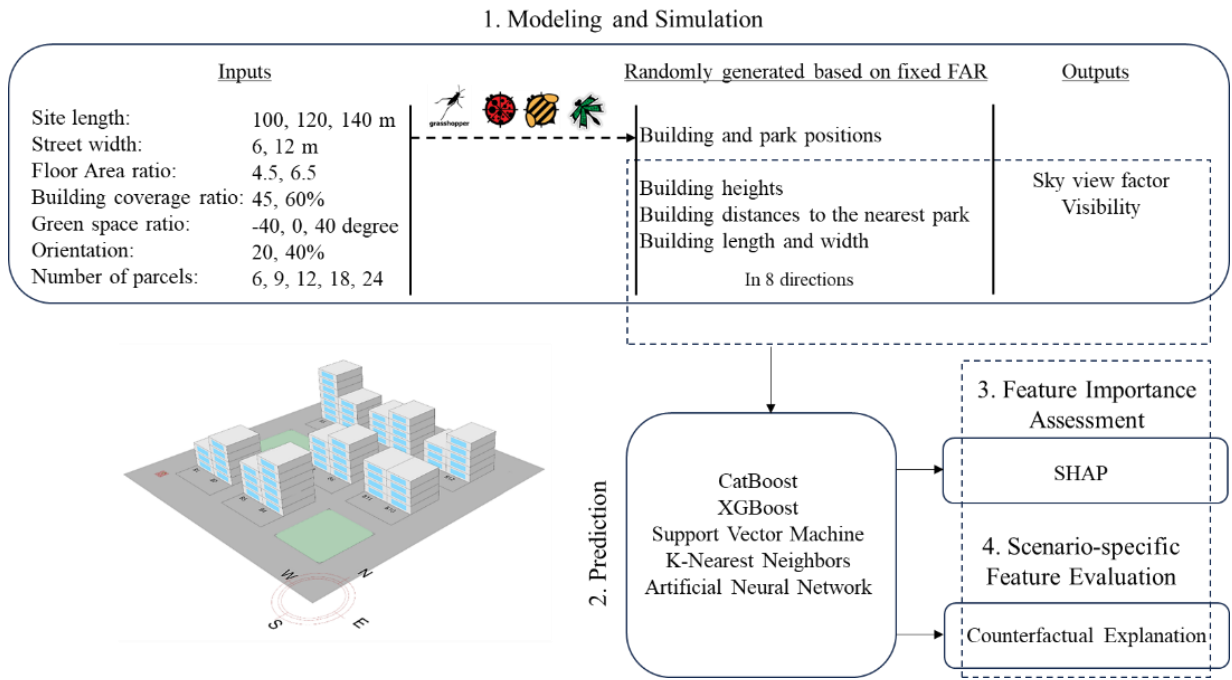

Figure 1. Methodological framework.

### 3.1. Development of urban setting scenarios

Due to variations in building heights and street widths, a series of urban blocks were developed following the typical configuration observed in Tehran, Iran, using Rhino, Grasshopper, and plugins like Ladybug, Dragonfly, and Honeybee 1.6.0. The modeling, performed on a regular grid, featured buildings ranging from 3 to 10 floors, each 3.5 meters high. The minimum urban block area was set at 1 hectare (100x100 meters) with street widths of 6 and 12 meters, consistent with Tehran's master plan. Divisions along the X and Y axes generated blocks with 6 to 24 parcels, each containing two buildings or a park. Building coverage ratios were set at 60% and 45%, with a fixed WWR of 40% on both north and south facades. Orientations were set at 40, -40, and 0 degrees, aligning with Tehran's layout, and green space ratios were 20% and 40%. In addition, the research was conducted with a constant Floor Area Ratio (FAR) of 4.5 and 6.5.

After modeling each urban block, the heights and distances of neighboring buildings surrounding parks in eight directions were recorded to analyze their individual effects on SVF and visibility. By treating height and distance as separate variables, the study aimed to isolate their distinct impacts on environmental performance, allowing for a more precise understanding of how each factor influences SVF and visibility independently. While these variables could interact, separating them provides valuable insights into their unique contributions, which might be obscured if analyzed together. To solely assess the impact of building adjacencies, it was assumed that there were no trees or vegetation influencing SVF or visibility. This controlled approach aligns with the hypothesis of this study, which focuses exclusively on the spatial arrangement of built structures. Incorporating vegetation would add complexity, increasing the number of solutions and potentially complicating the algorithm's testing process. With these defined input variables, the tool generated 1152 urban blocks randomly.

Regarding the outputs, SVF was computed using the Ladybug view percent component, which emits rays from observation points towards the sky to determine the proportion of sky visible without obstructions. For this study, the park parcel was divided into 1x1 meter grids to determine the average SVF for the entire surface. Park visibility was measured from two window points on each building using Ladybug Visibility Percent. All data were stored in Excel using Colibri and TT Toolbox plug-ins.

### 3.2. Prediction models

After modeling and preparing the dataset, the study evaluates five MLMs—CatBoost, XGBoost (Extreme Gradient Boosting), Support Vector Machine (SVM), K-Nearest Neighbors (KNN), and Artificial Neural Network (ANN)—to predict spatial metrics. These algorithms were selected for their diverse strengths in predictive modeling. CatBoost(Banik and Biswas, 2023) handles categorical features efficiently through ordered boosting. XGBoost(Mohammadiziazi and Bilec, 2020) is optimized for speed and performance using decision tree ensembles. KNN(Singaravel, 2020) classifies data based on proximity to neighboring points, making it effective for certain datasets. SVM(Drucker *et al.*, 1997) excels at defining hyperplanes for complex decision boundaries, while ANN (Ali *et al.*, 2020) is well-suited for learning intricate patterns in highly complex tasks.

SVF, which measures sky openness from specific points, is critical for environmental and climatic assessments. Its continuous nature makes regression modeling ideal for capturing variations, providing detailed insights into sunlight exposure and thermal comfort. In contrast, visibility influences human perception and space utilization, making classification more suitable for intuitive analysis. Based on simulation results, visibility is categorized into three levels—Level 0 (30–50%), Level 1 (50–75%), and Level 2 (75–100%)—to enhance interpretability and decision-making. This dual approach (1) Aligns each metric's analysis with urban planning needs, maximizing relevance and utility, (2) Improves result clarity, ensuring outputs are accessible for practical applications, and (3) Utilizes SHAP and CFXs across regression and classification models to analyze feature importance and performance.

Following model selection, the dataset is split into 80% training and 20% testing to validate accuracy and performance. Table 1 outlines the MLMs and their evaluation parameters.

Table 1. MLMs investigated, parameters and values employed.

| Algorithm | Hyperparameter | Value | Algorithm | Hyperparameter | Value |
|---|---|---|---|---|---|
| CatBoost | learning_rate | 0.05 | XGBoost | learning_rate | 0.1 |
| | depth | 6 | | max_depth | 5 |
| | iterations | 100 | | n_estimators | 50 |
| | l2_leaf_reg | 3 | | subsample | 0.8 |
| | colsample_bylevel | 0.8 | | colsample_bytree | 0.9 |
| kNN | n_neighbors | 5 | SVM | C | 1.0 |
| | weights | 'distance' | | kernel | 'rbf' |
| | p | 2 (Euclidean distance) | | gamma | 'scale' |
| ANN | learning_rate | 0.01 | | | |
| | hidden_layer_sizes | (100, 50) | | | |
| | activation | 'relu' | | | |
| | batch_size | 32 | | | |
| | epochs | 10 | | | |

**Evaluation indicators:** Evaluation metrics are essential for assessing the performance of MLMs in predicting target outcomes. In this study, MLMs are evaluated based on their ability to predict visibility (classification) and SVF (regression) accurately. For visibility, Accuracy and F1-Score are used: Accuracy measures the proportion of correct predictions, indicating overall model effectiveness, and F1-Score combines precision and recall, offering a nuanced view by balancing the model's ability to correctly identify relevant instances. For SVF, regression metrics include $R^2$, a measure of the explained variance; MSE (Mean Squared Error), which measures the average squared difference between predicted and actual values; and MAE (Mean Absolute Error), which computes the average absolute difference between predicted and actual values.

### 3.3. Significance of features

SHAP is used to investigate relationships between input features and target variables, providing interpretable insights by estimating feature contributions. The complexity of MLMs, often seen as "black boxes," poses challenges for domain experts during decision-making process (Doshi-Velez and Kim, 2017). The lack of transparency in these models can undermine users' ability to trust and understand the results (Chen *et al.*, 2023). To address this, post hoc explanation techniques like SHAP are applied for interpretability and transparency.

SHAP, based on game theory, calculates Shapley values to assess feature contributions and interactions, offering a comprehensive perspective on their importance (Slack *et al.*, 2020). It is preferred over traditional methods like gain or split count due to its fairness, stability, and ability to explain individual predictions. By ranking feature contributions clearly, SHAP reveals patterns in complex models and aids experts in making informed decisions (Markarian *et al.*, 2024). SHAP calculates feature importance using the definitions shown in Equation 1 (Nourkojouri *et al.*, 2021):

$$\phi_i(p) = \sum_{S \subseteq N/i} \left( \frac{|S|!(n-|S|-1)^i}{n!} \left[ p\left(S \cup \text{i}\right) - p(S) \right) \right) \tag{1}$$

- N: A set containing n features

- S: A subset excluding the specific feature i for which $\phi_i$ is being determined

- S∪{i}: A subset of features that includes the features in S plus the feature i

- S⊆N\{i}: All possible subsets S derived from the complete set N, excluding the feature i

Unlike SHAP, which explains feature contributions, CFXs methodology using the KD (KD-Tree) optimization approach is designed to efficiently generate minimal, actionable changes in input parameters to achieve specific objectives (Alfeo *et al.*, 2023), such as increasing SVF or enhancing visibility. The process begins by defining the objective and organizing the parameter space (e.g., building height, distance, orientation) using a KD-tree, which divides the space to quickly search for and locate the closest feasible point that meets the objective, minimizing the computational time. The algorithm evaluates which features are most influential, prioritizing these to achieve the desired counterfactual outcome. By leveraging the efficiency of the KD-tree, it avoids exhaustive searching and focuses on finding the nearest valid scenario that adheres to all constraints, such as zoning regulations or orientations.

Once a feasible scenario is identified, the algorithm refines it to ensure minimal and practical adjustments, such as small increases in building height or slight distance changes. The KD-tree structure allows rapid evaluation until the optimal solution is found, providing detailed recommendations and showing the impact of changes. This approach is particularly effective for scenarios that require rapid evaluation and adaptation of urban configurations, offering a significant efficiency advantage over iterative methods like GA.

In this study, CFXs are tested on both classification and regression models to ensure robustness across different prediction scenarios. Together, SHAP and CFXs enhance model interpretability, supporting transparent decision-making and practical guidelines for optimizing SVF and visibility in urban design.

### 3.4. Model Validation and Computational Efficiency Analysis

This study uses GA optimization to benchmark calculation time and validate the efficiency of the proposed framework. This method employed a computationally efficient process, starting with a randomly generated set of values (gene population) combined with corresponding tilts, enabling extensive search capabilities (Saad and Araji, 2021). The GA is configured with a population size of 50 and default settings for other evolutionary parameters, including maximum stagnation (50 generations), initial boost (2), maintain rate (5%), and inbreeding control (75%).

GA optimizes SVF and visibility metrics using single-objective optimization, providing a baseline for evaluating computational performance and solution quality. The results are compared with the CFXs approach, which suggests targeted modifications to building configurations, achieving faster and more interpretable optimizations. Additionally, Grasshopper simulations recreate scenarios based on the CFX recommendations, validating their feasibility and accuracy in real-world urban configurations.

## 4. Results and discussion

This section presents the study's findings in four parts. First, model evaluation results for SVF and visibility are discussed, comparing MLM performance using key metrics. Next, feature importance is assessed with SHAP to interpret how spatial attributes influence SVF and visibility, visualized through detailed plots highlighting key configurations like building height and distance. CFXs are then employed to provide targeted recommendations for improving these metrics, with scenarios analyzed for effectiveness by showing how minimal building adjustments achieve desired outcomes. Finally, computational efficiency and validation of CFX suggestions are evaluated. The GA benchmarks optimization time for calculating SVF and visibility, while Grasshopper simulations verify CFX recommendations by comparing predicted and simulated outcomes, confirming the methodology's reliability and practicality in realistic urban settings.

## 4.1. Model evaluation

The model evaluation for SVF and visibility was conducted using five different MLMs to identify the most accurate predictors for each metric. The evaluation focuses on key performance indicators such as $R^2$, MSE, and MAE for regression models, and F1-scores for classification models.

**SVF:** Table 2 presents evaluation indicators of the five regression models for predicting SVF. Analyzing the $R^2$ values, which gauge goodness of fit, CatBoostRegressor and XGBRegressor exhibit high scores of 0.903 and 0.908, respectively, indicating they capture substantial variance in the dependent variable. Lower MSE and MAE values are preferred for predictive accuracy. CatBoostRegressor and XGBRegressor also display the lowest MSE (164.75, 156.52) and MAE (6.3, 5.99) values. In contrast, SVR shows weaker performance across all metrics, with the lowest $R^2$ and higher MSE and MAE, indicating lower predictive capability. Overall, CatBoostRegressor and XGBRegressor emerge as the leading models for accurate regression predictions in the SVF dataset.

Table 2. Performance metrics of MLMs used for predicting the SVF.

| Model | $R^2$ | MSE | MAE |
|---|---|---|---|
| **KNeighborsRegressor** | 0.72 | 476.4 | 11.63 |
| **CatBoostRegressor** | 0.90 | 164.7 | 6.32 |
| **XGBRegressor** | 0.91 | 156.5 | 5.99 |
| **MLPRegressor** | 0.76 | 413.8 | 14.51 |
| **SVR** | 0.13 | 1478.3 | 26.6 |

**Visibility:** When evaluating MLMs for the three-class classification task, it is crucial to consider their performance across individual classes (0, 1, 2). As shown in Table 3, XGBoost and CatBoost consistently demonstrate high performance across all classes. For class 0, XGBoost achieves an F1-score of 0.8, while CatBoost achieves 0.78. In class 1, both models exhibit higher F1-scores, with XGBoost and CatBoost at 0.89. For class 2, XGBoost and CatBoost maintain robust F1-scores of 0.87 and 0.88, respectively. Comparatively, KNN and SVC show lower F1-scores across all classes. Despite comparable performance, this study ultimately employed XGBoost.

Table 3. Performance metrics of MLMs used for predicting visibility.

| Model | F1-score | | | Accuracy |
|---|---|---|---|---|
| | 0 | 1 | 2 | |
| **XGBoost** | 0.8 | 0.89 | 0.87 | 0.85 |
| **KNN** | 0.58 | 0.77 | 0.74 | 0.7 |
| **CatBoost** | 0.78 | 0.89 | 0.88 | 0.85 |
| **ANN** | 0.79 | 0.85 | 0.83 | 0.81 |
| **SVC** | 0.24 | 0.65 | 0.60 | 0.53 |

In summary, the evaluation of MLMs for predicting SVF and visibility highlights that XGBRegressor and CatBoostRegressor are the most accurate models for regression tasks related to SVF, demonstrating the highest $R^2$ values and the lowest MSE and MAE scores. For visibility classification, XGBoost and CatBoost consistently achieve high F1-scores across all classes, confirming their robustness. Despite similar performance levels, XGBoost was chosen for its slightly higher accuracy and consistency in results.

## 4.2.    Feature Importance Assessment

The SHAP value graph illustrates the influence of different features on the model's output, helping to understand each feature's contribution to the prediction. Features with larger SHAP values and wider distributions have a more significant impact. The SHAP values also provide insights into how the spatial placement of buildings or structures relative to the park can influence SVF and visibility within the park.

**SHAP analysis for SVF:** The SHAP plot analysis for SVF in Figure 2 highlights the multifaceted influences of urban features. The broad distribution of the park area indicates its crucial role in increasing SVF, with larger parks generally contributing to a more SVF. The dual impact of building width reflects its complex relationship with SVF, suggesting that both narrow and wide buildings can affect sky openness differently, potentially due to their placement and spatial configuration.

The length of buildings around the park tends to decrease SVF, suggesting that elongated structures might block more sky, thus reducing SVF. For adjacent building heights, the east and west-side building heights are particularly impactful, as increased height tends to lower SFV. The reduced variability in the influence of building distances in the southwest and south on SVF suggests these factors have a nuanced impact. In these directions, proximity may not consistently affect SVF, possibly due to varying urban layouts and building densities.

The height of buildings located on the north side inversely correlates with SVF. Similarly, increasing the distance between the park and buildings on the east and west sides, considering the characteristics of other buildings, can either increase or decrease the SVF. However, reducing the distance in these directions results in a decrease in SVF. Changes in street width also have an inverse relationship with this output. Other factors have less impact compared to those mentioned.

The last input with the least impact is orientation, indicating that buildings oriented towards the east have a lower SVF.

The circular bar charts in Figure 2 illustrate the SHAP values for building height and distance relative to SVF in eight directions, with different colors corresponding to specific SHAP value ranges. These values show how the distances and heights of buildings in various directions from a park influence the SVF perceived from within the park.

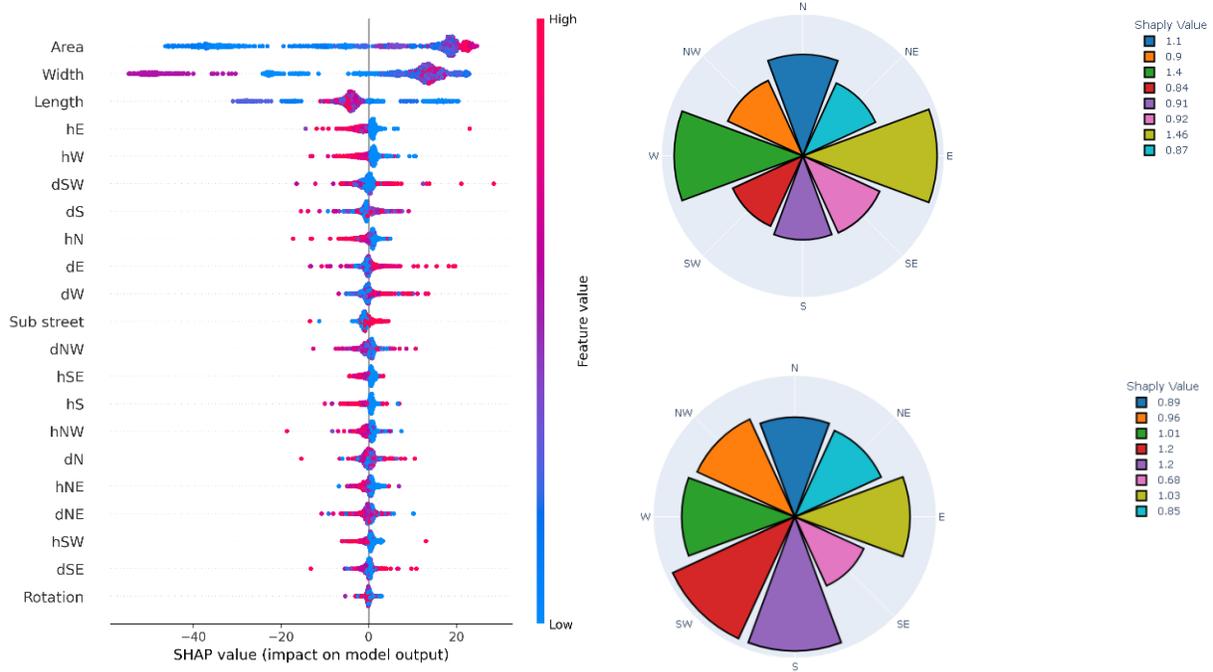

Figure 2 SHAP plot analysis (Left) and SHAP values for building height (up) and distance (down) for SVF.

For distances, the SHAP values indicate the most significant impact on SVF comes from the south (dS) and southwest (dSW) distances, each with a value of 1.2. The east (dE) and west (dW) building distances follow with SHAP values of 1.03 and 1.01, respectively, and the northwest building distance (dNW) has a value of 0.96. The north (dN) building distance has a SHAP value of 0.89, while the northeast (dNE) distance has a moderately lower impact. The southeast distance (dSE) displays the least influence with a SHAP value of 0.68.

For building heights, the SHAP values show that east building height (hE) has the highest impact on SVF with a SHAP value of 1.46, followed closely by west building height (hW) with a value of 1.4. The north building height (hN) also shows a substantial impact with a value over 1. The southeast (hSE), south (hS), and northwest buildings height (hNW) have lower SHAP values at 0.92, 0.91, and 0.9, respectively. The northeast (hNE) and southwest buildings height (hSW) have the least influence on SVF, with SHAP values below 0.9.

**SHAP analysis for visibility:** In the SHAP plot for visibility (Figure 3), several key features emerge as significant contributors to the predictive model's outcomes. The distance from the southern building stands out as the most impactful feature, showcasing a broad distribution of data points. This suggests that variations in the distance to the southern building significantly influence visibility outcomes, with data points spread across both positive and negative SHAP values, indicating that both shorter and longer distances can affect visibility in various ways.

The width of the buildings also exhibits a pronounced impact on visibility, with a notable concentration of data points highlighting its significance. The higher density of red points underscores the importance of building width in determining visibility, suggesting that reducing the width of buildings can enhance visibility. Additionally, factors such as the length of surrounding buildings and their distances from the park, particularly in specific directions like the north and northwest, also demonstrate notable impacts on visibility. This highlights the complex interplay of various factors in shaping visibility outcomes in urban settings.

After building width, the distance to the southeast building is recognized as the second most influential factor on visibility. Altering this distance generally enhances visibility, with decreasing the distance proving to be more impactful than increasing it. While the park area also plays a role in visibility outcomes, its impact is notably lesser compared to the SVF plot. Larger park areas tend to correlate with improved visibility to some extent, but the influence is not as pronounced as other factors.

Building orientation, though significant, ranks in the middle compared to other features. The distribution of SHAP values suggests that certain orientations, like east-facing (-40 degrees), may have a more positive impact on visibility, highlighting the importance of optimizing building placement and orientation. Analyzing building heights reveals that the height of the east-side building inversely affects visibility the most. Following this, the height of the southern building shows varying effects under different conditions. Conversely, increasing the height of the northwest building leads to increased visibility. Other components have a lesser impact compared to those mentioned.

As depicted in Figure 3, the SHAP values for building heights indicate that the east building height (hE) has the most significant impact on park visibility, with a value of 0.18, suggesting that structures to the east of the parks are particularly influential. Following this, the south (hS), northwest (hNW), and southeast (hSW) buildings height directions have SHAP values of 0.13, 0.12, and 0.11, respectively. The west (hW) and southeast (hSE) buildings height have equal impacts with a value of 0.1, followed by the northeast building height (hNE). The north building height (hN) has the least impact, with a value of 0.08, indicating that heights in this direction are less critical in affecting park visibility.

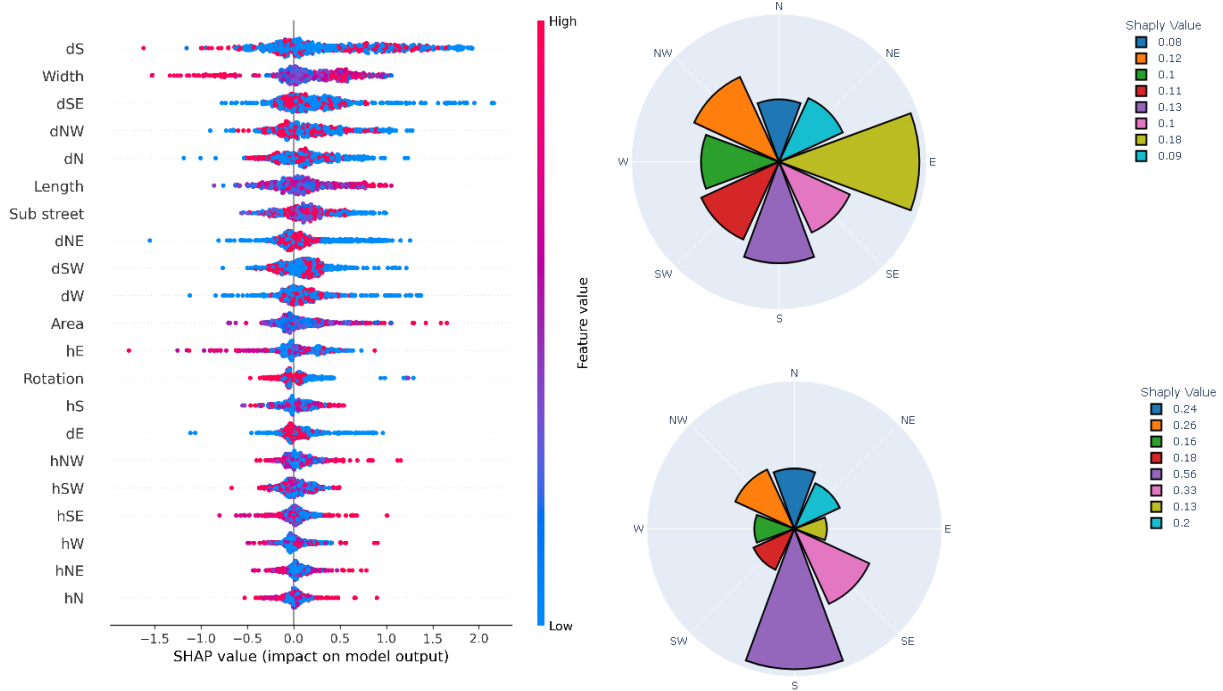

Figure 3 SHAP plot analysis (Left) and SHAP values for building height (up) and distance (down) for Visibility.

The SHAP values for building distances indicate that the distance of buildings in the south (dS) direction has the highest impact on park visibility, with a value of 0.56. The distances of buildings in the northwest (dNW) and southeast (dSE) directions also have considerable impacts, with values of 0.26 and 0.33, respectively. The northeast (dNE) and north (dN) distances have moderate influences, while the southwest (dSW), west (dW), and east (dE) distances have the lowest impacts on visibility, with SHAP values ranging from 0.11 to 0.2. Although these distances do impact park visibility, their influence is comparatively weaker than that of buildings in the south and southeast directions.

In conclusion, the SHAP analysis provides a detailed understanding of the spatial features influencing SVF and visibility within urban configurations. For SVF, park area and building heights (especially on the east and west sides) are the most influential factors, indicating that increasing park size and managing building height can significantly enhance sky openness. The impact of building distance varies depending on the direction, demonstrating the nuanced effects of spatial arrangements. For visibility, the SHAP values highlight the importance of the distance from the southern building and building width, suggesting that optimizing these elements can improve visibility outcomes. Although park area and building heights in other directions also affect visibility, their impact is comparatively less significant.

These insights confirm that strategic adjustments in building placement, orientation, and height can optimize SVF and visibility, providing urban planners with actionable guidelines for designing functional and visually open urban spaces.

### 4.3. Counterfactual Explanations

The utilization of CFXs begins with establishing specific urban configurations for optimization. This research generated 10 urban configurations, and for each configuration, 5 strategies were developed using CFXs, focusing separately on optimizing SVF and visibility. Out of these, 2 scenarios are highlighted in Tables 4 and 5, showcasing the strategies devised for each configuration. These scenarios vary in urban block dimensions, orientations, and other characteristics, providing a comprehensive evaluation of CFX effectiveness.

The SVF scenario includes seven buildings surrounding a 292 square meter park, oriented at 40 degrees north. Each building measures 16 by 11 meters, resulting in an SVF of 68%. The visibility scenario features a site oriented at -40 degrees north with five buildings, each 12.5 by 17 meters, facilitating a 73% visibility to the park (Class 1). The street width remains constant at 6 meters in both scenarios. The next step is to define the optimization goals: increasing the SVF percentage and elevating the visibility class by one level. Figure 4 shows the main scenarios for SVF and visibility. Targeted space is the park at the center of the each block.

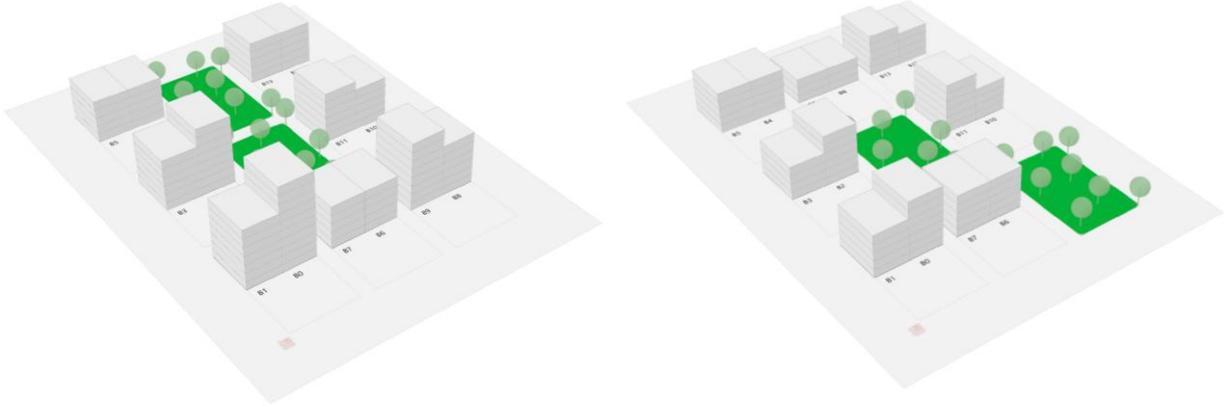

Figure 4 Main scenarios block configuration for SVF (left) and Visibility (right).

Humans often consider alternative outcomes by modifying controllable, action-oriented aspects. However, CFXs might reveal that some key factors are beyond control, rendering them non-actionable. For instance, suggesting changes in building orientation to increase SVF may be impractical due to the immobility of urban blocks. Actionable CFXs are valuable for feasible alterations, like adjusting building heights. Identifying non-actionable features can expose biases, highlighting the importance of context and fairness in decision systems. In this study, neighborhood characteristics such as orientation, street width, building width, length, and park area were held constant. CFXs offer strategies based on defined scenarios, achieving outcomes with minimal adjustments. Five distinct strategies were devised for each output, providing insights into urban planning, optimizing visibility while preserving other neighborhood characteristics. Tables 4 and 5 detail the neighborhood characteristics and changes for each variable per strategy. A zero indicates no alteration to that feature.

**CFX for SVF:** Table 4 presents a comparative analysis between the original characteristics of an urban block and different strategies (S1 to S5) proposed by CFX models to achieve specific SVF goals. Each strategy provides solutions for altering building heights and distances to enhance the SVF of the urban block. The "main" column displays the baseline measurements, such as the number of stories (h) in each direction and the distances (d) from the park (in meters). The strategy columns show the proposed changes by the models, while the last row indicates the resulting changes in SVF for each strategy.

Table 4. Main scenario and proposed strategies by CFX for SVF enhancements.

| | Main | S1 | S2 | S3 | S4 | S5 |
|---|---|---|---|---|---|---|
| hN | 0 | | 0 | | 0 | |
| hNE | 5 | | | | | |
| hE | 5 | 0 | 4 | | -5 | 0 |
| hSE | 8 | | | | | |
| hS | 5 | | 0 | | 0 | |
| hSW | 10 | | | | | |
| hW | 5 | -2 | -4 | 0 | -6 | -7 |
| hNW | 5 | 4 | | | | |
| dN | 33 | | | | | 0 |
| dNE | 21 | | | | | |
| dE | 24 | | | | | 36 |
| dSE | 32 | 0 | 0 | | 0 | |
| dS | 18 | | | 22 | | |
| dSW | 13 | | | | | 0 |
| dW | 25 | | 0 | | | |
| dNW | 22 | | | | | |
| SVF | 48 | 2 | 9 | 4 | 12 | 13 |

For example, in S1, reducing the height of the western building by 2 stories and increasing the height of the northwest building by 4 stories leads to a 2% increase in SVF. In S2, increasing the east building height by 4 stories and reducing the west building height by 4 stories results in a 9% increase in SVF, raising it from 68% to 77%. In S3, a significant distance change is proposed for the south (dS) by 22 meters, raising the SVF to 72%. Similarly, in S4, reducing the height of the buildings on the east and west sides by 5 and 6 stories, respectively, enhances the SVF by 12% to 80%. Lastly, S5 involves changing both the distance and height of the east and west-side buildings by 36 meters and 7 stories, respectively, increasing the SVF by 13%.

**CFX for visibility:** These changes in visibility are aimed at upgrading the visibility class from 0 to 1 across all strategies (Table 5). Notably, reducing the height of buildings in the east (hE) consistently emerges as an effective decision, highlighting its importance in enhancing visibility. This aligns with insights from the SHAP values and circular bar charts. Regarding changes in building distances from the park, scenarios exhibit diverse alterations, particularly in the southeast (dSE) direction. Both increasing and decreasing distances from the park in this direction consistently improve visibility, reaffirming its significance in influencing visibility outcomes.

Table 5. Main scenario and proposed strategies by CFX for visibility enhancements

| | Main | S1 | S2 | S3 | S4 | S5 |
|---|---|---|---|---|---|---|
| hN | 3 | 0 | 5 | 0 | 0 | 0 |
| hNE | 5 | | 2 | | | |
| hE | 6 | -2 | -1 | | -2 | |
| hSE | 0 | 0 | 0 | | 0 | |
| hS | 5 | | | | | |
| hSW | 7 | -3 | -4 | | | -3 |
| hW | 6 | 0 | 0 | | | 0 |
| hNW | 5 | | | | | |
| dN | 23 | | | -14 | | |
| dNE | 41 | | | 18 | | |
| dE | 31 | | | 0 | | |
| dSE | 41 | | | -19 | | |
| dS | 5 | | | 17 | | |
| dSW | 18 | | | 15 | 24.0 | 24 |
| dW | 27 | | | 0 | 0 | 0 |
| dNW | 20 | | | | | 3 |
| Visibility | 0 | 1 | 1 | 1 | 1 | 1 |

In Strategy 1 (S1), minor adjustments in building heights at the east and southwest, with no changes in distances, lead to improved visibility. S2 suggests selective increases in heights in the north (hN) direction and decreases in the east (hE) and southwest (hSW) directions. S3 proposes notable alterations in distances of dN, dNE, dSE, dS, and dSW with northeast building height changes, contributing to changes in the visibility class. S4 involves reducing the height in the east direction and increasing the distance of buildings in the southwest direction, leading to substantial improvements. Similarly, S5 recommends a 3-story reduction for the southwestren building, and an increase in the southeast and northwest buildings distances by 14 and 3 meters, respectively, resulting in visibility enhancements.

### 4.4.    Evaluation of CFX Strategies Using Genetic Optimization

**GA Evaluation:** The GA was used to benchmark the optimization time in comparison to the CFXs approach for improving SVF and visibility across the urban configurations. The objective was to assess which method achieves the desired results in the shortest time while maintaining accuracy.

The time comparison between the GA and CFX approaches reveals significant differences in computational efficiency. For the GA, each iteration for SVF scenarios takes approximately 6 seconds, accounting for the calculation time and adjustments made during the process. Visibility scenarios take longer, with each iteration averaging 9 seconds due to the additional complexity of factors such as the WWR. It is important to note that the initial iterations in the GA do not necessarily yield optimized results, as the algorithm explores a wide range of possibilities before gradually converging towards an optimal solution. In the tested GA case, with a population size of 50 and no time limit, the closest results for SVF were reached in the 3rd generation, taking about 15 minutes, while visibility required 4 generations, taking approximately 30 minutes to approach the optimal value.

In contrast, the tested CFX approach is considerably faster, with the number of suggested counterfactuals set at 5, making the total time around 1 minute to provide the results. This makes CFX approximately 15 times faster than GA for SVF and 30 times faster for visibility. While the time per counterfactual in CFX might appear higher than a single GA iteration, it is important to understand that CFX does not engage in exhaustive searching. Instead, it provides optimized solutions directly within this timeframe. Although GA iterations may take less time individually, the initial iterations are not optimized and do not follow the principle of minimal adjustments. Additionally, each GA optimization must be customized for the specific problem at hand, which can add to its complexity and setup time. Overall, CFX offers a much quicker and more efficient solution, particularly for scenarios requiring rapid evaluation and precise adjustments in urban configurations.

**Simulation Validation:** Grasshopper simulations were conducted to validate the CFX recommendations for each of the 10 urban configurations. The accuracy of the CFX outputs was evaluated using the Root Mean Square Error (RMSE) between the predicted and simulated SVF and visibility values (Figure 5). The box plot analysis reveals key insights into the RMSE values across the 10 urban configurations. The minimum RMSE values range from 1.19% to 2.65%, indicating excellent accuracy in those scenarios, while the maximum values range from 5.72% to 9.73%, all remaining below the 10% threshold, demonstrating overall good accuracy even in the most extreme cases. The mean RMSE values across configurations fall between 4.12% and 6.06%, showing that the CFX predictions generally maintain acceptable accuracy levels. Median RMSE values, consistent with the mean, range from 2.91% to 6.41%, confirming low error margins in most scenarios.

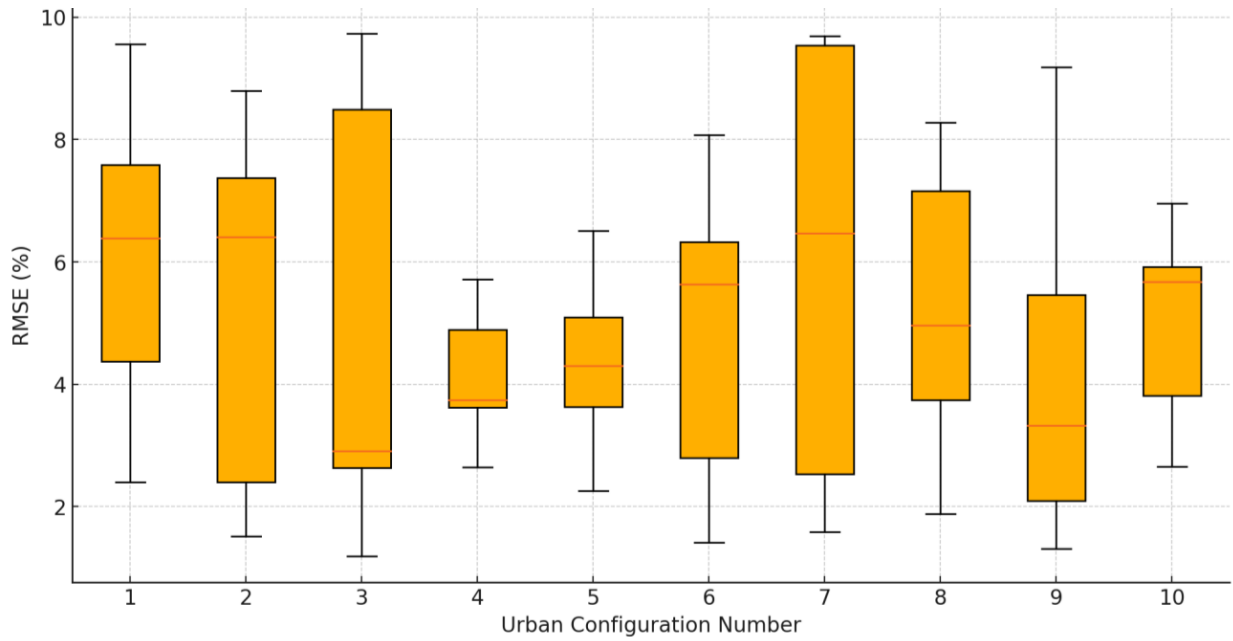

Figure 5 RMSE values across urban configurations scenarios.

The distribution of RMSE values reveals that, in most cases, the deviations between the CFX recommendations and the Grasshopper simulations are minimal, indicating that the CFX strategies are generally accurate. However, the RMSE values revealed slight deviations in certain scenarios, particularly where large adjustments in building heights or distances were recommended. The overall low RMSE values suggest that the CFX recommendations are feasible and accurate, but further refinement may be needed for configurations with extreme parameter adjustments to minimize discrepancies.

## 5. Conclusion

Open spaces are essential for urban quality of life, providing areas for recreation, social interaction, and environmental benefits. Key metrics like SVF and visibility influence thermal comfort, aesthetics, safety, and the usability of these spaces. These two metrics strongly rely on the configuration of neighboring buildings and the overall urban fabric, making their optimization challenging due to the complex interactions between built and natural elements. This study aimed to address this challenge by developing advanced predictive models and methodologies to accurately predict and optimize SVF and visibility in urban contexts. Traditional methods often lack the efficiency and flexibility needed to account for the intricate relationships between urban features, prompting the need for data-driven solutions that can provide practical insights for urban planning.

XGBoost was identified as the most effective model for predicting SVF and visibility, demonstrating strong performance across evaluation metrics. Feature importance analysis using SHAP values revealed that factors like park area, building width, and the heights of west and east buildings are critical for SVF, while building width and the distance to southern buildings are influential for visibility, emphasizing the reliance of these metrics on the configuration of neighboring buildings and the urban layout. SHAP values also offered transparency and interpretability, allowing urban planners to understand the impact of each feature on the outcomes, making the process more informed and data-driven. To optimize these metrics efficiently, this study proposes a CFX approach that provides minimal-change strategies, significantly outperforming traditional GA. While the GA required 15 minutes for SVF and 30 minutes for visibility with a population size of 50, the CFX approach achieved similar outcomes in just 1 minute. Grasshopper simulations validated the accuracy and practicality of the CFX solutions, demonstrating their applicability in real-world urban planning.

In conclusion, this research showcases the effectiveness of integrating advanced modeling techniques—XGBoost, SHAP, and CFXs—into urban planning, optimizing SVF and visibility to enhance open spaces. Future research should explore simultaneous optimization of multiple metrics and include natural elements like trees for a more comprehensive analysis of urban configurations.